\documentclass[runningheads]{llncs}
\usepackage[T1]{fontenc}
\usepackage{graphicx}
\usepackage{amssymb}
\usepackage{amstext}
\usepackage{booktabs}
\usepackage{multirow}
\usepackage{caption}
\usepackage{bbding}

\begin{document}
	\title{Beyond Adapting SAM: Towards End-to-End Ultrasound Image Segmentation via Auto Prompting}
	\author{Xian Lin\inst{1} %index{Lin, Xian}
		\and
		Yangyang Xiang\inst{1} %index{Xiang, Yangyang}
		\and
		Li Yu\inst{1} %index{Yu, Li}
		\and
		Zengqiang Yan\inst{1}\textsuperscript{(}\Envelope\textsuperscript{)} %index{Yan, Zengqiang}
	}
	\authorrunning{X. Lin et al.}
	\institute{School of Electronic Information and Communications, Huazhong University of Science and Technology \\
		\email{\{xianlin, xyy\_2001, hustlyu, z\_yan\}@hust.edu.cn}}
	\maketitle              % typeset the header of the contribution
	\begin{abstract}
		
		End-to-end medical image segmentation is of great value for computer-aided diagnosis dominated by task-specific models, usually suffering from poor generalization. With recent breakthroughs brought by the segment anything model (SAM) for universal image segmentation, extensive efforts have been made to adapt SAM for medical imaging but still encounter two major issues: 1) severe performance degradation and limited generalization without proper adaptation, and 2) semi-automatic segmentation relying on accurate manual prompts for interaction. In this work, we propose SAMUS as a universal model tailored for ultrasound image segmentation and further enable it to work in an end-to-end manner denoted as AutoSAMUS. Specifically, in SAMUS, a parallel CNN branch is introduced to supplement local information through cross-branch attention, and a feature adapter and a position adapter are jointly used to adapt SAM from natural to ultrasound domains while reducing training complexity. AutoSAMUS is realized by introducing an auto prompt generator (APG) to replace the manual prompt encoder of SAMUS to automatically generate prompt embeddings. A comprehensive ultrasound dataset, comprising about 30k images and 69k masks and covering six object categories, is collected for verification. Extensive comparison experiments demonstrate the superiority of SAMUS and AutoSAMUS against the state-of-the-art task-specific and SAM-based foundation models. We believe the auto-prompted SAM-based model has the potential to become a new paradigm for end-to-end medical image segmentation and deserves more exploration. Code and data are available at \url{https://github.com/xianlin7/SAMUS}.
		
	\end{abstract}
	\section{Introduction}
	
	Medical image segmentation, a crucial technology to discern and highlight specific organs, tissues, and lesions within medical images, serves as an integral component of computer-aided diagnosis systems~\cite{liu2021review}. Numerous deep-learning models have been proposed for medical image segmentation, showcasing substantial potential~\cite{unet,missformer}. However, these models are tailored for specific objects and necessitate training new model parameters when applied to other objects, resulting in great inconvenience for clinical applications with diverse tasks.
	
	Segment anything model (SAM), serving as a versatile foundation model for vision segmentation, has garnered considerable acclaim owing to its remarkable segmentation capabilities across diverse objects and robust zero-shot generalization capacity~\cite{sam}. According to user prompts, including points, bounding boxes, and coarse masks, SAM is capable of segmenting the corresponding objects. Therefore, through simple prompting, SAM can be effortlessly adapted to various segmentation applications. This paradigm enables the integration of multiple individual medical image segmentation tasks into a unified framework (\textit{i.e.}, a universal model), greatly facilitating clinical deployment~\cite{samsurvey}.
	
	Despite constructing the largest dataset to date (\textit{i.e.}, SA-1B), SAM encounters a rapid performance degradation in the medical domain due to the scarcity of reliable clinical annotations~\cite{samsurvey}. Some foundation models have been proposed to adapt SAM to medical image segmentation by tuning SAM on medical datasets~\cite{medsam,msa}. However, the same as SAM, they perform a no-overlap 16× tokenization on the input images before feature modeling, which destroys the local information crucial for identifying small targets and boundaries, making them struggle to segment clinical objects with complex/threadlike shapes, weak boundaries, small sizes, or low contrast. In addition, these SAM-based models require manually providing task-related prompts to generate the corresponding masks, leading to a semi-automatic segmentation pipeline. Such a paradigm is inflexible when dealing with certain clinical tasks.
	
	In this paper, we present SAMUS first to transfer the strong feature representation ability of SAM to the domain of medical image segmentation, and then extend the trained SAMUS into an automatic version (\textit{i.e.}, AutoSAMUS) to flexibly handle various downstream segmentation tasks. Specifically, SAMUS inherits the ViT image encoder, prompt encoder, and mask decoder of SAM, with tailored designs to the image encoder. First, we shorten the sequence length of the ViT branch by reducing the required input size to lower the computational complexity. Then, a feature adapter and a position adapter are developed to fine-tune the ViT image encoder from natural to medical domains. To complement local (\textit{i.e.}, low-level) information in the ViT image encoder, we introduce a parallel CNN-branch image encoder, running alongside the ViT-branch and propose a cross-branch attention module to enable each patch in the ViT-branch to assimilate local information from the CNN-branch. A large ultrasound dataset called US30K is constructed to comprehensively train and evaluate the efficacy of SAMUS. After obtaining the trained SAMUS, it is expected to run in an end-to-end manner for downstream specific tasks. Therefore, we extend SAMUS into AutoSAMUS by introducing an auto prompt generator (APG) with learnable task tokens to replace the manual prompt encoder of SAMUS for generating task-related prompt embeddings. Experimental results demonstrate that SAMUS outperforms the state-of-the-art (SOTA) task-specific and universal segmentation approaches. More importantly, based on the trained SAMUS, adjusting AutoSAMUS on specific tasks can realize end-to-end automatic segmentation and achieve better segmentation performance compared to SOTA task-specific methods. This indicates that developing auto-prompted SAM-based models is promising as a new end-to-end segmentation paradigm.
	
	\section{Related Works}
	\label{sec:relatedwork}
	\textbf{Adapt SAM to Medical Image Segmentation.} SAM has demonstrated remarkable performance in natural images but struggles with some medical image segmentation tasks, especially on objects with complex shapes, blurred boundaries, small sizes, or low contrast~\cite{samsurvey}. To bridge this gap and enable SAM to adapt effectively to the medical image domain, several methods have been proposed by applying vision tuning techniques to SAM. Specifically, MedSAM trains SAM on medical images by freezing the prompt encoder, focusing on tuning the image encoder and mask decoder~\cite{medsam}. SAMed applies the low-rank-based (LoRA) strategy on the image encoder to tune SAM at a lower computational cost, making it more feasible for medical image segmentation~\cite{samed}. MSA adopts down-ReLU-up adapters on the ViT image encoder and mask decoder to introduce medical information~\cite{msa}. Compared to current SAM-based universal models, the proposed SAMUS focuses more on complementing local features and realizing end-to-end automatic segmentation.
	
\noindent \textbf{Prompts in SAMs.} Vanilla SAM generates task-related masks under the driven of precise spatial prompts, \textit{e.g.}, points, bounding boxes, and masks. To automatically obtain these spatial prompts, some methods introduce separate input-related networks. Specifically, AdapterShadow, SAC, and UV-SAM use Efficient-Net, U-Net, and SegFormer respectively to generate coarse masks for making spatial prompts~\cite{adaptershadow,SAC,uvsam}. The introduced parameters of such separate networks are on the same level as task-specific methods, making the universal model cumbersome. Polyp-SAM++ uses Grounding DINO to generate bounding box prompts from the text prompts~\cite{polypsam}. Adaptive SAM and SP-SAM encode text prompts into prompt embeddings by CLIP~\cite{adaptivesam,spsam}. Although these approaches can effectively utilize text information, there is a lack of text-image data in medical scenes to tune the text encoders trained on natural scenes. SurgicalSAM proposes a prototype-based class prompt encoder to generate the dense and sparse prompt embeddings~\cite{surgicalsam}. Auto-prompting SAM develops an auto-prompt encoder by constructing Up-Down full convolution layers~\cite{autopromptsam}. These prompt encoders are deeply coupled with the mask decoder, causing difficulty in constructing robust feature representations through multi-objective learning for SAM-based models. Comparatively, the proposed APG is a lightweight and independent module and highly extendable to other SAM-based foundation models.
	
	\section{Method}
	\label{sec3}
	
	\begin{figure}[t]
		\centering
		\includegraphics[width=1\columnwidth]{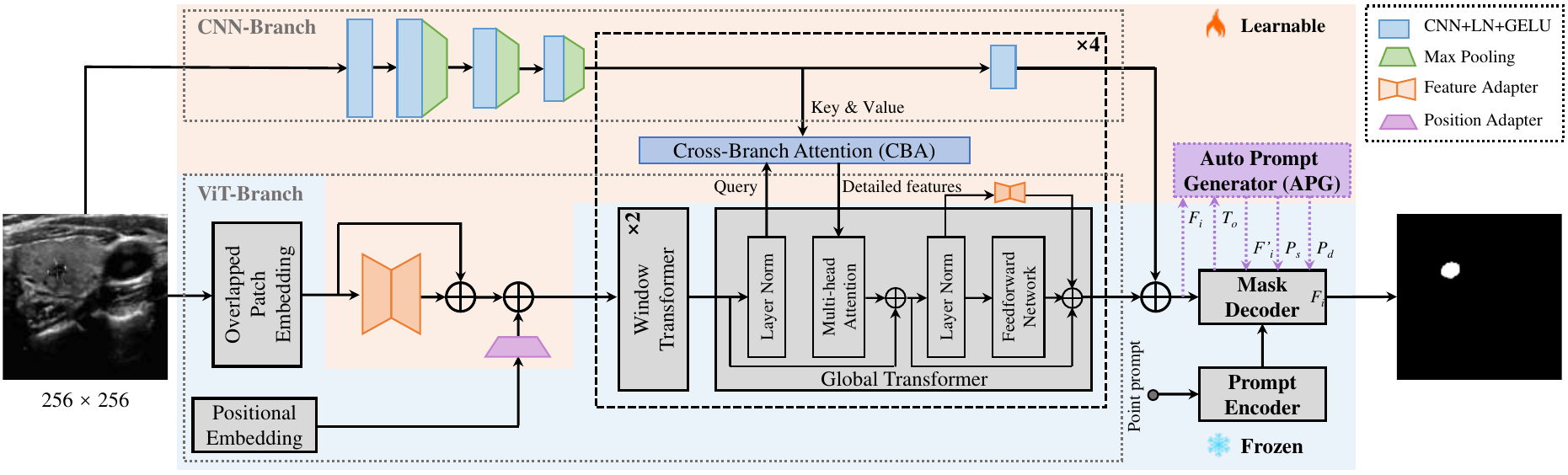}
		\caption{Overview of the proposed SAMUS. APG represents the expansion module for extending SAMUS into AutoSAMUS.}
		\label{fig1}
	\end{figure}
	
	\textbf{The Overall Architecture of SAMUS.} As depicted in Fig.~\ref{fig1}, the overall architecture of SAMUS is inherited from SAM, retaining the structure and parameters of the prompt encoder and the mask decoder without any adjustment. Comparatively, the image encoder is carefully modified to address the challenges of inadequate local features and excessive computational memory consumption, making it more clinically friendly. Major modifications include reducing the required input size, overlapping the patch embedding, introducing adapters to the ViT branch, adding a CNN branch, and introducing cross-branch attention (CBA). Specifically, the input spatial resolution is scaled down from $1024\times1024$ pixels to $256\times256$ pixels, resulting in a substantial reduction in GPU memory cost due to the shorter input sequence in transformers. The overlapped patch embedding uses the same parameters as the patch embedding in SAM while its patch stride is half to the original stride, well keeping the information from patch boundaries. Adapters in the ViT branch include a position adapter and five feature adapters. The position adapter is to accommodate the global position embedding in shorter sequences due to the smaller input size. The first feature adapter follows the overlapped patch embedding to align input features with the required feature distribution of the pre-trained ViT image encoder. The remaining feature adapters are attached to the residual connections of the feed-forward network in the global transformer to fine-tune the pre-trained image encoder. In terms of the CNN branch, it is parallel to the ViT branch, providing complementary local information to the latter through the CBA module, which takes the ViT-branch features as the query and builds global dependency with features from the CNN branch. It should be noted that CBA is only integrated into each global transformer. Finally, the outputs of both the two branches are combined as the final encoded image embedding $F_i$ of SAMUS.
	
	\noindent \textbf{Adapters in the ViT Branch.}
	To facilitate the generalization of the trained image encoder (\textit{i.e.}, the ViT branch) of SAM to smaller input sizes and the medical image domain, we introduce a position adapter and five feature adapters. These adapters can effectively tune the ViT branch while only requiring much fewer parameters. Specifically, the position adapter is responsible for adjusting the positional embedding to match the resolution of the embedded sequence. It begins by downsampling the positional embedding through max pooling with the stride and kernel size as 2, achieving the same resolution as the embedded sequence. Then, a convolution operation with a kernel size of $3\times3$ is applied to tune the position embedding, further aiding the ViT branch in better handling smaller inputs. All feature adapters have the same structure that comprises three components: a down linear projection, an activation function, and an up linear projection. The procedure of each feature adapter can be formulated as:
	\begin{equation}
		\mathcal{A}(x) = \mathcal{G}(xE_d)E_u,
		\label{eq1}
	\end{equation}
	where $\mathcal{G}$ represents the GELU activation founction, $E_d \in \mathbb{R} ^ {d \times \frac{d}{4}}$ and $E_u \in \mathbb{R} ^ {\frac{d}{4} \times d}$ are the projection matrices, $d$ is the dimension of the feature embedding. Through these simple operations, feature adapters enable the ViT branch to better adapt to the feature distribution of medical image domains.
	
	\begin{figure}[!t]
		\centering
		\includegraphics[width=1\columnwidth]{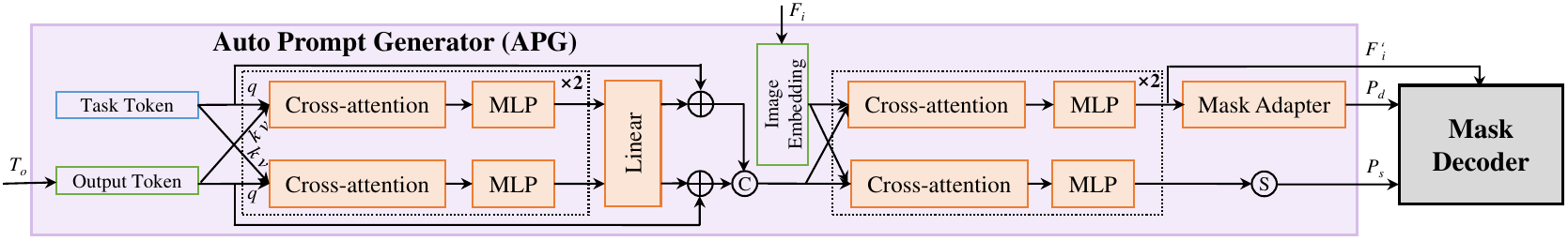}
		\caption{Details of the auto prompt generator.}
		\label{fig2}
	\end{figure}
	
	\noindent \textbf{The CNN Branch.} The CNN branch consists of a series of convolution-pooling blocks connected in sequence. Specifically, the inputs pass through a single convolution block initially, followed by being processed through three convolution-pooling blocks. Then, the feature maps in the CNN branch share the same spatial resolution as those of the ViT branch. In the rest part of the CNN branch, such single convolution blocks are repeated four times in sequence. Each single convolution block contains a convolution with a kernel size of $3\times3$ and each convolution-pooling block contains a max pooling with the stride and kernel size as 2 and a single convolution block. More details are illustrated in Fig.~\ref{fig1}. This minimalist and lightweight design of the CNN branch is to prevent overfitting during training.
	
	\noindent \textbf{Cross-branch Attention.}
	The cross-branch attention (CBA) module creates a bridge between the CNN branch and the ViT branch to further complement missing local features with the ViT branch. For a pair of feature maps from the ViT branch $F_v$ and the CNN branch $F_c$, cross-branch attention in the single head can be formulated as:
	\begin{equation}
		\mathcal{F}(F_v,F_c) = (\sigma(\frac{F_vE_q(F_cE_k)^T}{\sqrt{d_m}}) + R)(F_cE_v),
		\label{eq2}
	\end{equation}
	where $\sigma$ represents the Softmax function. $E_q \in \mathbb{R} ^ {d \times d_m}$, $E_k \in \mathbb{R} ^ {d \times d_m}$, and $E_v \in \mathbb{R} ^ {d \times d_m}$ are the learnable weight matrices used to project $F_c$ and $F_v$ to different feature subspaces. $R \in \mathbb{R} ^ {hw \times hw}$ is the relative position embedding, and $d_m$ is the dimension of CBA. The final output of CBA is the linear combination of $g$ such single-head attention.
	
	\noindent \textbf{AutoSAMUS.} AutoSAMUS, an end-to-end automatic segmentation framework extended from SAMUS, is realized by replacing the manual prompt encoder of SAMUS with APG. As depicted in Fig.~\ref{fig2}, the inputs of APG consist of the output tokens $T_o \in \mathbb{R} ^ {5 \times d}$ and the image embedding $F_i$, where $T_o$ is the frozen parameters extracted from the mask decoder and $F_i$ is the output of the image encoder. To indicate the segmentation task, APG introduces learnable task tokens $T_t \in \mathbb{R} ^ {k \times d}$ for automatically generating task-related prompt embeddings, where $k$ is the number of task tokens. Firstly, the combination of cross-attention and multi-layer perceptron (MLP) is used to couple the task tokens and the output tokens, formulated as:
	\begin{equation}
		\mathcal{C}(T_t,T_o) = \textbf{MLP}(\sigma(\frac{T_tW_q(T_oW_k)^T}{\sqrt{d}})(T_oW_v)),
		\label{eq3}
	\end{equation}
	where \textbf{MLP} consists of two linear layers. $W_q$, $W_k$, and $W_v \in \mathbb{R} ^ {d \times d}$ are the learnable weight matrices. Then, the updated task tokens are represented as:
	\begin{equation}
		T_{t_1} = \mathcal{C}(\mathcal{C}(T_t,T_o), \mathcal{C}(T_o,T_t))W + T_t,
		\label{eq4}
	\end{equation}
	where $W\in \mathbb{R} ^ {d \times d}$ is the projection matrix. Similarly, by swapping the positions of $T_t$ and $T_o$ in Eq.~\ref{eq4}, the updated output tokens $T_{o_1}$ can be calculated. Next, to adapt the image embedding to the task domain and make the task tokens aware of image information, we perform the combination operation $\mathcal{C}$ defined in Eq.~\ref{eq3} between the image embedding and the combined tokens $T = [T_{t_1}, T_{o_1}]$. After that, the updated image embedding and the combined tokens are represented as $F'_i = \mathcal{C}(\mathcal{C}(F_i,T), \mathcal{C}(T,F_i))$ and $T' = \mathcal{C}(\mathcal{C}(T,F_i), \mathcal{C}(F_i,T))$. Finally, based on $T'$ and $F'_i$, APG generates the sparse prompt embedding $P_s = T'[:k,:]$ and the dense prompt embedding $P_d = \mathcal{M}(F'_i)$ to prompt the frozen mask decoder, where $\mathcal{M}$ represents the sequence of operation consisting of four single convolution blocks with channels of $\frac{d}{4}$, $\frac{d}{4}$, $\frac{d}{4}$, and $d$. In addition, the updated image embedding $F'_i$ will replace the original image embedding to participate in mask decoding.
	
	\begin{table}[t]
		\centering
		\caption{Universality comparison of SAMUS and SOTA foundation models on segmenting thyroid nodule (TN3K), breast cancer (BUSI), left ventricle (CAMUS-LV), myocardium (CAMUS-MYO), and left atrium (CAMUS-LA).}\label{tab1}
		%\resizebox{\columnwidth}{!}{
			\begin{tabular}{c|cc|cc|cc|cc|cc}
				\hline
				& \multicolumn{2}{c|}{TN3K}       & \multicolumn{2}{c|}{BUSI}       & \multicolumn{2}{c|}{CAMUS-LV}   & \multicolumn{2}{c|}{CAMUS-MYO}                & \multicolumn{2}{c}{CAMUS-LA}                                                  \\
				\multirow{-2}{*}{Method} & Dice           & HD             & Dice           & HD             & Dice           & HD             & Dice           & HD                           & Dice                                  & HD                                    \\ \hline
				SAM~\cite{sam}                      & 29.59          & 134.87         & 54.01          & 82.39          & 28.18          & 196.64         & 29.42          & 184.10                       & 17.28                                 & 193.70                                \\
				MedSAM~\cite{medsam}                   & 71.09          & 42.91          & 77.75          & 34.26          & 87.52          & 15.28          & 76.07          & 25.72                        & 88.06                                 & 15.70                                 \\
				SAMed~\cite{samed}                    & 80.40          & 31.29          & 74.82          & 34.60          & 87.67          & 13.24          & 82.60          & 19.48 & 90.92                                 & 12.60                                 \\
				MSA~\cite{msa}                      & 82.67          & 29.15          & 81.66          & 28.87          & 90.95          & \textbf{11.29} & 82.47          & 19.28                        & 91.80                                 &  \textbf{11.59}\\
				SAMUS                    & \textbf{83.05} & \textbf{28.82} & \textbf{84.54} & \textbf{27.24} & \textbf{91.13} & 11.76          & \textbf{83.11} & \textbf{18.99}               &\textbf{92.00} & 12.08                                 \\ \hline
			\end{tabular}
			%}
	\end{table}
	
	\section{Experiments}
	\textbf{Datasets.}
	To comprehensively evaluate the effectiveness of SAMUS, we have constructed a large ultrasonic dataset named US30K as summarized in the supplementary material, containing data from seven publicly-available datasets, including TN3K~\cite{tn3k}, DDTI~\cite{ddti}, TG3K~\cite{tg3k}, BUSI~\cite{busi}, UDIAT~\cite{udiat}, CAMUS~\cite{camus}, and HMC-QU~\cite{hmcqu}. The data in TN3K and TG3K is partitioned into train, validation, and test sets following TRFE~\cite{tn3k}. BUSI is randomly split into 7:1:2 for training, validation, and testing, respectively. CAMUS is divided into a train set and a test set first according to the challenge~\cite{camus}. Then, we randomly select 10$\%$ patients from the train set to validate the model and the rest data as the final training data. To evaluate the generalization of different models, the other datasets in US30K are unseen during training. The comparison between SAMUS and other foundation models is conducted by training on the entire training set of US30K and evaluated on separate tasks. For a fair comparison, all foundation models are re-implemented and trained for 400 epochs under the same settings using the same single-point prompt. The comparison between SAMUS and SOTA task-specific approaches is conducted by training on each single dataset.
	
		\begin{table}[t]
		\centering
		\caption{Generalization comparison of foundation models on segmenting thyroid nodule (DDTI), breast cancer (UDIAT), and myocardium (HMC-QU)}\label{tab2}
		%\resizebox{\columnwidth}{!}{
		\begin{tabular}{c|cc|cc|cc|cc|cc}
			\hline
			\multirow{2}{*}{Dataset} & \multicolumn{2}{c|}{SAM~\cite{sam}} & \multicolumn{2}{c|}{MedSAM~\cite{medsam}} & \multicolumn{2}{c|}{SAMed~\cite{samed}} & \multicolumn{2}{c|}{MSA~\cite{msa}} & \multicolumn{2}{c}{SAMUS}       \\
			& Dice       & HD          & Dice         & HD           & Dice    & HD               & Dice   & HD              & Dice           & HD             \\ \hline
			DDTI                     & 25.57      & 116.23      & 57.94        & 51.77        & 61.64   & \textbf{43.45}   & 62.24  & 46.49           & \textbf{66.78} & 44.35          \\
			UDIAT                    & 49.18      & 104.43      & 61.49        & 50.80         & 72.56   & 30.82            & 76.24  & \textbf{26.64}  & \textbf{78.06} & 26.91          \\
			HMC-QU                    & 25.91      & 93.20        & 34.58        & 36.30         & 37.82   & 37.72            & 40.56  & 38.18           & \textbf{56.77} & \textbf{25.21} \\ \hline
		\end{tabular}
		%}
	\end{table}
	
	\begin{figure}[!t]
		\centering
		\includegraphics[width=1\columnwidth]{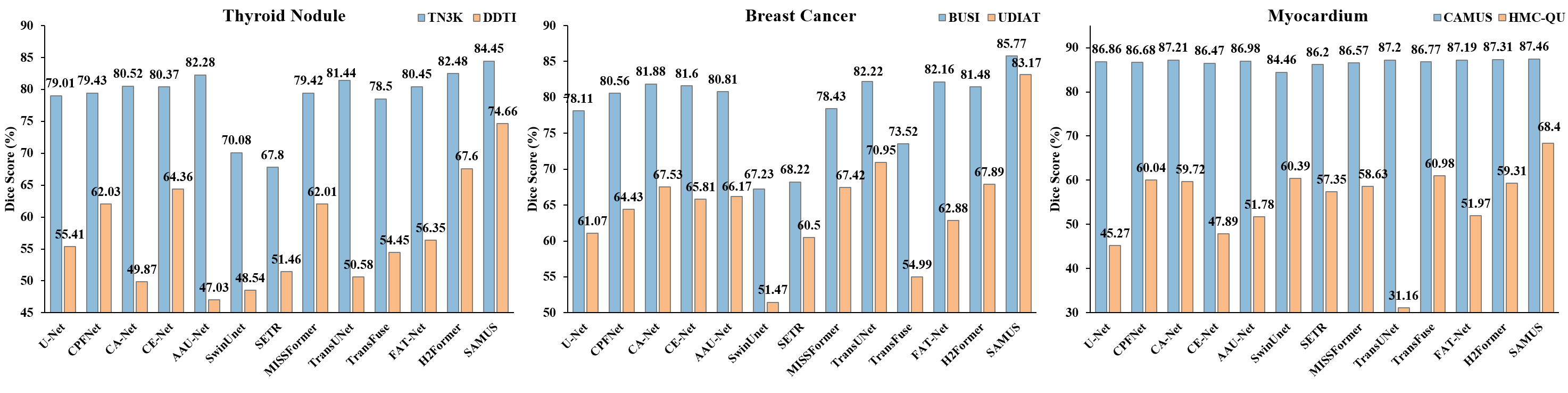}
		\caption{Comparison between SAMUS and task-specific methods evaluated on seeable (marked in blue) and unseen datasets (marked in orange).}
		\label{fig3}
	\end{figure}
	
	\noindent \textbf{Compare SAMUS with SOTA Foundation Models.}
	Comparison of universality and generalization among foundation models on US30K are summarized in Tables~\ref{tab1} and \ref{tab2}. Among comparison foundation models, MAS~\cite{msa} is the best-performing model both in universality and generalization. Compared to MSA, SAMUS consistently achieves remarkable improvements across all subdatasets of US30K. It indicates the value of the CNN branch and the CBA module in SAMUS for complementing local information which is crucial for medical image segmentation.
	
	\begin{table}[!t]
		\centering
		\caption{Quantitative dice ($\%$) comparison of AutoSAMUS and SOTA task-specific methods. AutoSAMUS- is the model when only APG is updated.}\label{tab3}
		\resizebox{\columnwidth}{!}{
			\begin{tabular}{c|ccccccc|cc}
				\hline
				Method & U-Net & CPFNet & CANet & AAU-Net & MISSFormer & TransUNet & H2Former & AutoSAMUS- & AutoSAMUS \\ \hline
				DDTI   & 76.71 & 78.38  & 75.79 & 78.16  & 78.11      & 77.52     & 77.92    & 78.89      & \textbf{82.64}     \\
				UDIAT  & 82.64 & 82.37  & 82.68 & 81.46  & 82.89      & 81.15     & 82.26    & 84.17      & \textbf{85.39}     \\
				HMC-QU  & 93.26 & 93.58  & 93.65 & 93.66  & 93.50       & 93.45     & 93.44    & 92.87      & \textbf{94.10}     \\ \hline
			\end{tabular}
		}
	\end{table}
	
	\begin{table}[!t]
		\centering
		\caption{Ablation study on different component combinations of SAMUS. F-Adapter and P-Adapter represent feature and position adapters respectively.}\label{tab4}
		\resizebox{\columnwidth}{!}{
			\begin{tabular}{cccc|cc|cc|cc|cc}
				\hline
				\multicolumn{4}{c|}{Components}                & \multicolumn{2}{c|}{TN3K}                               & \multicolumn{2}{c|}{DDTI}                               & \multicolumn{2}{c|}{BUSI}                              & \multicolumn{2}{c}{UDIAT}                              \\
				CNN Branch & CBA       & F-Adapter & P-Adapter & Dice                      & HD                          & Dice                      & HD                          & Dice                      & HD                         & Dice                      & HD                         \\ \hline
				\XSolidBrush     & \XSolidBrush    & \XSolidBrush    & \XSolidBrush    & \multicolumn{1}{l}{29.59} & \multicolumn{1}{l|}{134.87} & \multicolumn{1}{l}{25.57} & \multicolumn{1}{l|}{116.23} & \multicolumn{1}{l}{54.01} & \multicolumn{1}{l|}{82.39} & \multicolumn{1}{l}{49.18} & \multicolumn{1}{l}{104.43} \\
				\Checkmark  & \XSolidBrush    & \XSolidBrush    & \XSolidBrush    & 82.17                     & 31.41                       & 68.31                     & 48.66                       & 81.42                     & 29.50                      & 82.24                     & 22.53                      \\
				\Checkmark  & \Checkmark & \XSolidBrush    & \XSolidBrush    & 83.65                     & 28.47                       & 72.71                     & 35.76                       & 83.53                     & 30.26                      & 80.87                     & 25.60                      \\
				\XSolidBrush     & \XSolidBrush    & \Checkmark & \XSolidBrush    & 83.64                     & 29.83                       & 70.38                     & 45.29                       & 84.53                     & 26.30                      & 81.25                     & 23.18                      \\
				\XSolidBrush     & \XSolidBrush    & \XSolidBrush    & \Checkmark & 80.19                     & 32.12                       & 63.67                     & 53.86                       & 80.78                     & 29.00                      & 79.72                     & 24.71                      \\
				\Checkmark  & \Checkmark & \Checkmark & \Checkmark & \textbf{84.45}            & \textbf{28.22}              & \textbf{74.66}            & \textbf{21.03}              & \textbf{85.77}            & \textbf{25.49}             & \textbf{83.17}            & \textbf{21.25}             \\ \hline
			\end{tabular}
		}
	\end{table}

	\noindent \textbf{Compare SAMUS with SOTA Task-Specific Methods.}
	As depicted in Fig.~\ref{fig3}, 12 SOTA methods are included for comparison~\cite{unet,missformer,cpfnet,canet,cenet,aaunet,swinunet,setr,transunet,transfuse,fatnet,h2former}. SAMUS surpasses the best comparison methods across all datasets including TN3K, BUSI, CAMUS-MYO, DDTI, UDIAT and HMC-QU with an average increase of 1.97$\%$, 3.55$\%$, 0.15$\%$, 7.06$\%$, 12.22$\%$ and 7.42$\%$ in Dice, respectively. It proves the necessity of adapting SAM to the medical image domain by SAMUS. 
	
	\noindent \textbf{Compare AutoSAMUS with SOTA Task-Specific Methods.} To compare AutoSAMUS with task-specific methods, we first load the trained parameters of SAMUS into AutoSAMUS. Then, we fine-tune APG and APG together with the learnable parts of SAMUS on three downstream tasks, the results are represented by AutoSAMUS- and AutoSAMUS respectively in Table~\ref{tab3}. AutoSAMUS- surpasses the best comparison method on DDTI and UDIAT and approaches the best comparison method on HMC-QU, while AutoSAMUS outperforms the best comparison method on DDTI, UDIAT, and HMC-QU with an average increase of 4.26$\%$, 2.5$\%$, and 0.44$\%$ in Dice, respectively.
	
	\noindent \textbf{Effectiveness of each component in SAMUS.}
	As summarized in Table~\ref{tab4}, coupling any component of SAMUS can effectively improve the segmentation performance and generalization ability of SAM. Combining all four components, SAMUS achieves the best segmentation and generalization performance.
8	
	\section{Conclusion}
	
	In this paper, we propose SAMUS, a universal foundation model derived from SAM, and its end-to-end version AutoSAMUS, for clinically-friendly and generalizable ultrasound image segmentation. Specifically, we present a CNN branch image encoder, a feature adapter, a position adapter, and a cross-branch attention module to enrich the feature representations of ultrasound objects. Furthermore, to facilitate the clinical application of downstream tasks, we combine SAMUS with an auto prompt generator for automatic segmentation, which realizes a new end-to-end segmentation paradigm instead of relying on manual prompts as vanilla SAM. A large ultrasound image dataset US30K consisting of 30k+ images and 68k+ masks is constructed for evaluation and potential clinical usage. Extensive experiments demonstrate the outstanding performance of SAMUS and AutoSAMUS, outperforming SOTA both SAM-based medical foundation models and task-specific models.
	
	\begin{credits}
		\subsubsection{\ackname}
		This work was supported in part by the National Natural Science Foundation of China under Grant 62271220 and Grant 62202179, in part by the Natural Science Foundation of Hubei Province of China under Grant 2022CFB585, and in part by the Fundamental Research Funds for the Central Universities, HUST: 2024JYCXJJ032. The computation is supported by the HPC Platform of HUST.
		\subsubsection{\discintname}
		The authors have no competing interests to declare that are relevant to the content of this article.
	\end{credits}

\end{document}